\documentclass{article}

\usepackage{PRIMEarxiv}

\usepackage{multirow}       % 合并行
\usepackage[utf8]{inputenc} % allow utf-8 input
\usepackage[T1]{fontenc}    % use 8-bit T1 fonts
\usepackage{hyperref}       % hyperlinks
\usepackage{url}            % simple URL typesetting
\usepackage{booktabs}       % professional-quality tables
\usepackage{amsfonts}       % blackboard math symbols
\usepackage{nicefrac}       % compact symbols for 1/2, etc.
\usepackage{microtype}      % microtypography
\usepackage{lipsum}
\usepackage{fancyhdr}       % header
\usepackage{graphicx}       % graphics
\graphicspath{{media/}}     % organize your images and other figures under media/ folder
\usepackage{graphicx}
\usepackage{float}
\usepackage{enumitem}

\title{TCM-3CEval: A Triaxial Benchmark for Assessing Responses From Large Language Models in Traditional Chinese Medicine}
\author{
  Tianai Huang\textsuperscript{1,†},
  Lu Lu\textsuperscript{2,†},
  Jiayuan Chen\textsuperscript{2},
  Lihao Liu\textsuperscript{2},
  Junjun He\textsuperscript{2},
  Yuping Zhao\textsuperscript{3}*,
  Wenchao Tang\textsuperscript{1}*,
  Jie Xu\textsuperscript{2}*  \\
  \textsuperscript{1}School of Acupuncture-Moxibustion and Tuina, Shanghai University of Traditional Chinese Medicine, Shanghai, China \\
  \textsuperscript{2}Shanghai Artificial Intelligence Laboratory, Shanghai, China \\
  \textsuperscript{3}China Academy of Chinese Medical Sciences, Beijing, China \\
  *Correspondence to: \texttt{\{Xu Jie\}xujie@pjlab.org.cn}, \texttt{\{Tang Wenchao\}vincent.tang@shutcm.edu.cn},\\ \texttt{\{Zhao Yuping\}18810084632@163.com}\\
  †These authors contributed equally.
}
\begin{document}
\maketitle

\begin{abstract}
% 待根据结果完善
Large language models (LLMs) excel in various NLP tasks and modern medicine, but their evaluation in traditional Chinese medicine (TCM) is underexplored. To address this, we introduce TCM-3CEval, a benchmark assessing LLMs in TCM across three dimensions: core knowledge mastery, classical text understanding, and clinical decision-making. We evaluate diverse models, including international (e.g., GPT-4o), Chinese (e.g., InternLM), and medical-specific (e.g., PLUSE). Results show a performance hierarchy: all models have limitations in specialized subdomains like Meridian \& Acupoint theory and Various TCM Schools, revealing gaps between current capabilities and clinical needs. Models with Chinese linguistic and cultural priors perform better in classical text interpretation and clinical reasoning. TCM-3CEval sets a standard for AI evaluation in TCM, offering insights for optimizing LLMs in culturally grounded medical domains. The benchmark is available on Medbench's TCM track, aiming to assess LLMs’ TCM capabilities in basic knowledge, classic texts, and clinical decision-making through multidimensional questions and real cases.
\end{abstract}

% keywords can be removed
\keywords{Benchmark \and Large language model \and Traditional Chinese medicine}

\section{Introduction}
With the rapid development of artificial intelligence, large language models (LLMs) are increasingly being applied in the medical field, particularly in Traditional Chinese Medicine (TCM). TCM LLMs are intelligent systems built on natural language processing (NLP)\cite{chowdhary2020natural} and deep learning technologies\cite{deng2014deep}. They aim to simulate the thinking process of TCM experts by learning from TCM literature, medical cases, and clinical data, and to provide diagnostic and therapeutic suggestions. Several TCM LLMs have been introduced\cite{dai2024tcmchat}, such as TCMLLM, ‘Shennong’\cite{dou2023shennonggpt}, ‘Zhongjing’\cite{yang2024zhongjing}, ‘Hua Tuo’\cite{wang2023huatuo}, TCM Pangu\cite{zeng2021pangu}, and ‘Dayi Jinkui’. These models leverage technologies like Recurrent Neural Networks (RNNs), Transformers, and Long Short-Term Memory networks (LSTMs)\cite{sherstinsky2020fundamentals}, combined with reinforcement learning, to develop medical reasoning capabilities for doctor-patient interactions and diagnostic systems. 

Compared with traditional medicine like TCM, the integration of modern medicine and LLM technology is more comprehensive and has developed more rapidly. For example, Med-PaLM 2, developed by Google, focuses on providing high-quality answers to medical questions and generating comprehensive summaries for medical professionals\cite{singhal2023expertlevelmedicalquestionanswering}. BioGPT, developed by Microsoft, excels in analyzing biomedical research and generating insightful summaries of medical literature\cite{Luo_2022}. They have shown promising results in various tasks, including question answering and text generation, making it a valuable tool for researchers and clinicians. In addition, the evaluation system for modern medical LLMs is more robust and comprehensive, typically focusing on the accuracy of data, consistency of diagnosis, and rationality of treatment plans, with a higher degree of standardization and quantifiability. For example, MedExQA is a benchmark test based on the United States Medical Licensing Examination (USMLE), used to assess models' performance in medical knowledge Q\&A\cite{kim2024medexqa}; PubMedQA focuses on Q\&A tasks related to biomedical literature\cite{jin2019pubmedqa}; and MedMCQA, developed based on Indian medical data, evaluates models' application capabilities in clinical practice\cite{pal2022medmcqa}. These benchmarks provide clear standards and quantifiable metrics for evaluating modern medicine models. \textit{However, in contrast to modern medicine models that rely on standardized and structured data, TCM LLMs face greater challenges due to the abstract nature of TCM theory and its highly individualized practice.}

The evaluation task of TCM LLMs is complex. TCM theory emphasizes the concepts of ‘Unity of Heaven and Man’ and ‘Syndrome Differentiation and Treatment’\cite{wang2019scientific, jiang2012syndrome}. The TCM theoretical system is intricate and abstract\cite{lozano2014basic}, and the diagnostic and treatment process is highly individualized. These characteristics demand a higher level of theoretical understanding and clinical reasoning ability for model evaluation. Current evaluation methods, such as the TCM Practitioner Qualification Examination and academic assessment schemes, have limitations. The examination focuses on testing basic theories and clinical knowledge through standardized questions but fails to comprehensively reflect practical application capabilities. Academic assessment schemes, such as the one proposed by East China Normal University, primarily focus on evaluating TCM LLMs from a technical perspective, but still require a multi-dimensional evaluation approach rooted in TCM disciplines that addresses the various branches of TCM knowledge\cite{yue2024tcmbench}. Moreover, existing modern medicine benchmarks (e.g., MedQA, MedMCQA) perform well in their respective fields but are insufficient for evaluating TCM models. Therefore, the ultimate goal of TCM LLMs is to assist in clinical diagnosis and treatment, yet \textit{existing evaluation methods do not fully reflect the models' decision-making capabilities in real clinical settings, resulting in a gap between evaluation outcomes and practical application needs.}

To address these limitations, we propose the ‘3C’ evaluation framework\autoref{fig:enter-label}, grounded in TCM talent cultivation standards and structured knowledge systems \cite{chen2013case, zhou2024integrating}. This framework assesses models across three critical dimensions:

\begin{figure}
    \centering
    \includegraphics[width=1.0\linewidth]{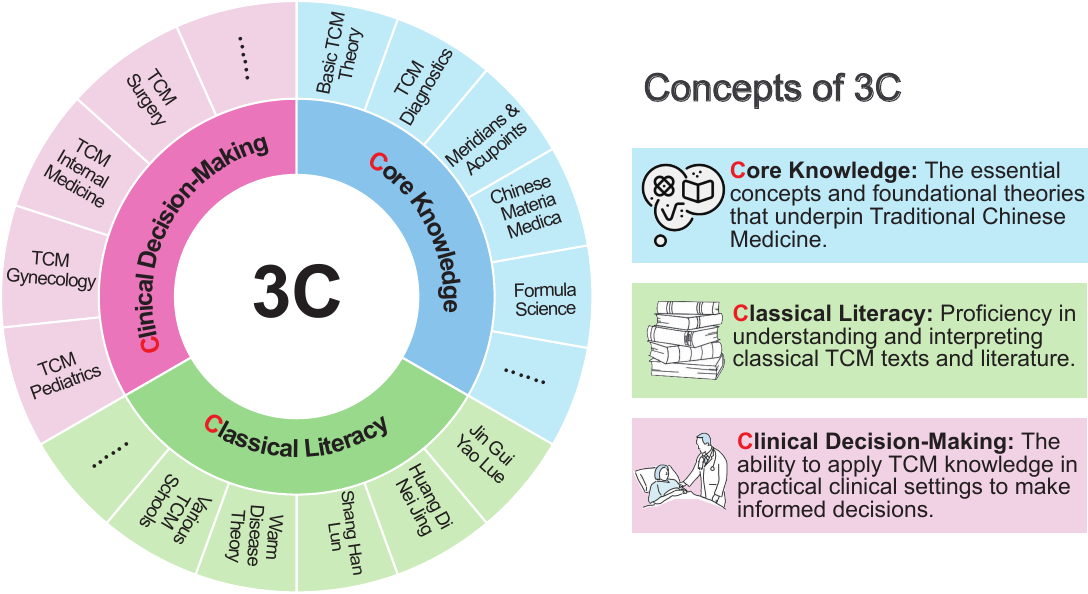}
    \caption{The concepts of ‘3C’}
    \label{fig:enter-label}
\end{figure}

\begin{itemize}[leftmargin=*]
\item[]  \textbf{(a) Core Knowledge}: Evaluates mastery of fundamental TCM concepts, including Yin-Yang theory, the Five Elements, viscera-meridian theory, and body fluids. These foundational principles are essential for accurate reasoning and decision-making.
    
\item[]  \textbf{(b) Classical Literacy}: Assesses the model’s comprehension of key TCM texts, such as the \textit{Huangdi Neijing} (Inner Canon of the Yellow Emperor) \cite{huangdi}, \textit{Shanghan Zabing Lun} (Treatise on Cold Damage and Miscellaneous Diseases) \cite{shanghan}, and \textit{Jin Kui Yao Lue} (Essential Prescriptions from the Golden Chamber) \cite{jinkui}. These texts provide theoretical and practical guidance for clinical practice.
    
\item[]  \textbf{(c) Clinical Decision-Making}: Examines the model’s ability to handle real-world clinical scenarios, including accurate syndrome differentiation, rational formula composition \cite{wang2021network}, and personalized treatment recommendations. For example, when diagnosing wind-cold syndrome, the model should recognize symptoms, propose suitable formulas like Mahuang Tang (Ephedra Decoction), and adjust treatments based on individual patient characteristics.
\end{itemize}

In summary, this study aims to establish a scientific and systematic evaluation framework for TCM LLMs to scientifically and objectively assess model performance. It will provide clear development goals for TCM LLMs, support TCM education, and enhance the intelligence of clinical diagnosis and treatment. Ultimately, by constructing a scientific evaluation system for TCM LLMs, we can promote the deep integration of TCM theory and modern technology, providing assistance for the modernization and internationalization of TCM.

\section{Methods}\label{sec11}
\subsection{Evaluation dimension design}\label{subsec2}
The evaluation framework of this paper revolves around the following three core dimensions, and show in Table \ref{tab:tcm_benchmark}:

\subsubsection{Assessment of Mastery of Basic TCM Knowledge}\label{subsubsec2}
This dimension aims to evaluate the model's grasp of the fundamental theories of TCM, covering four core courses: Basic TCM Theory, TCM Diagnostics, Chinese Materia Medica, and Formula Science. The assessment employs standardized exercise sets, including multiple-choice questions, fill-in-the-blank questions, and short-answer questions, to comprehensively examine the model's understanding of basic TCM concepts, principles, and methods.

\subsubsection{Assessment of Understanding of TCM Classics}\label{subsubsec2}

TCM classics are an important source of TCM theory and a core component of TCM learning. This dimension selects four major classics: 'Jin Gui Yao Lue' (Essential Prescriptions from the Golden Cabinet), 'Huang Di Nei Jing' (The Yellow Emperor's Classic of Internal Medicine), 'Shang Han Lun' (Treatise on Cold Damage and Miscellaneous Diseases), 'Warm Disease Theory', and Various TCM Schools. Exercise sets are designed to assess the model's ability to understand and interpret classical texts. Question types include analysis of classic texts and clinical application analysis.

\subsubsection{Assessment of TCM Clinical Decision-Making Ability}\label{subsubsec2}

This dimension focuses on the model's application ability in clinical practice, covering common diseases in internal medicine, surgery, gynecology, pediatrics, as well as TCM ophthalmology and otorhinolaryngology. This dimension selects exercise sets from 'TCM Surgery', 'TCM Internal Medicine', 'TCM Gynecology', and 'TCM Pediatrics' to evaluate the model's ability to propose diagnostic approaches, treatment plans, and medication recommendations based on patients' symptoms, signs, and medical history, thereby assessing its decision-making ability in actual clinical practice.

\begin{table}
\vspace{-27pt}
 \caption{TCM LLM Benchmark Dimensions}
  \centering
  \begin{tabular}{p{2cm}p{2.7cm}cp{6.3cm}}
    \toprule
    Dimension & Sub-dimension & Q\# & Main Evaluation Content \\
    \midrule
    \multirow{5}{*}{TCM-Exam} & Basic TCM Theory & 30 & Understanding of Yin-Yang theory, Five Elements, Zang-Fu organs, and fundamental principles of TCM pathophysiology. \\
    & TCM Diagnostics & 30 & Four diagnostic methods, pattern identification, tongue and pulse diagnosis techniques, and clinical examination procedures. \\
    & Meridians \& Acupoints & 30 & Knowledge of meridian pathways, acupoint locations, functions, contraindications, and therapeutic applications. \\
    & Chinese Materia Medica & 30 & Herb properties, flavors, channel tropism, therapeutic effects, dosage, contraindications, and common herb combinations. \\
    & Formula Science & 30 & Classical formulas, composition ratios, compatibility principles, modifications for different syndromes, and clinical applications. \\
    \midrule
    \multirow{5}{*}{TCM-LitQA} & Jin Gui Yao Lue & 30 & Interpretation of Zhang Zhongjing's "Essential Prescriptions of the Golden Cabinet" and its clinical significance. \\
    & Huang Di Nei Jing & 30 & Analysis of "The Yellow Emperor's Inner Classic" theoretical foundations, cosmology, and physiological concepts. \\
    & Shang Han Lun & 30 & Understanding of "Treatise on Cold Damage Disorders," six-stage pattern identification, and treatment strategies. \\
    & Warm Disease Theory & 30 & Comprehension of febrile disease patterns, four-level progression, treatment principles, and clinical manifestations. \\
    & Various TCM Schools & 30 & Knowledge of different theoretical schools, their founders, key contributions, and distinct clinical approaches. \\
    \midrule
    \multirow{4}{*}{TCM-MRCD} & TCM Surgery & 40 & Diagnosis and treatment of external conditions, abscesses, ulcers, wounds, and surgical interventions in TCM practice. \\
    & TCM Internal Medicine & 40 & Pattern differentiation and treatment of internal diseases affecting respiratory, digestive, and cardiovascular systems. \\
    & TCM Gynecology & 35 & Female-specific conditions, menstrual disorders, pregnancy-related issues, and their TCM-based treatments. \\
    & TCM Pediatrics & 35 & Child-specific conditions, growth and development patterns, pediatric diagnosis, and gentle treatment methods. \\
    \bottomrule
  \end{tabular}
  \label{tab:tcm_benchmark}
\vspace{-28pt}
\end{table}

\subsection{Evaluation Dataset Construction}\label{subsec2}

To ensure the scientific rigor and comprehensiveness of the evaluation, this study constructs a high-quality set of exercises based on TCM textbooks, classical literature, and clinical cases. Each dimension of the exercise set has been reviewed and revised by TCM experts to ensure the accuracy and representativeness of the content. The focus areas, characteristics, and application scenarios of the three types of datasets are shown in Table \ref{tab:dataset}.

\begin{table}[H]
    \centering
    \caption{Dataset Focus Areas and Characteristics}
    \begin{tabular}{p{2.8cm}p{3cm}p{3.2cm}p{3.2cm}}
    \toprule[1.5pt]
        Dataset Category & Focus Area & Characteristics & Application Scenarios\\
    \midrule[1pt]
        TCM Professional Knowledge Set & Basic theory, herbology, formulary & Broad coverage of knowledge points, diverse question types, theoretical focus & Evaluation of basic knowledge grasp and theoretical depth\\
        TCM Classics Exercise Set & Understanding and application of classical works & Focus on classical literature, emphasis on application, high difficulty & Assessment of classical theory application and complex reasoning\\
        Clinical Diagnosis and Treatment Case Set & Clinical diagnosis and treatment ability & Case-oriented, practical, comprehensive approach & Evaluation of clinical application ability and integrated skills\\
    \bottomrule[1.5pt]
    \end{tabular}
    \label{tab:dataset}
\end{table}

The specific composition of the datasets is as follows:

\textbf{1 TCM Professional Knowledge Exercise Set:} Covers core knowledge points from five courses: 'Formulary', 'Meridian and Acupoint Studies', 'Herbology', 'Basic Theory of TCM', and 'Diagnostics in TCM'. Each course includes 30 questions, totaling 150 questions. This dataset comprehensively evaluates the model's grasp of TCM theoretical knowledge through broad coverage of knowledge points and diverse question types. Its systematic and theoretical nature effectively tests the model's accuracy and depth in basic knowledge, providing foundational support for the comprehensive evaluation of the model's capabilities.

\textbf{2 TCM Classics Exercise Set}: Selects key chapters from five classical texts: 'Jin Gui Yao Lue', 'Nei Jing', 'Shang Han Lun', 'Wen Bing Xue', and 'Theories of Various TCM Schools'. Each text includes 30 questions, totaling 150 questions. This dataset not only tests the model's memory and understanding of the original texts but also emphasizes its ability to apply classical theories to practical problems. Due to the depth and complexity of classical works, this dataset is highly challenging and effectively evaluates the model's mastery of TCM classical theories and its reasoning and application abilities in complex scenarios.

\textbf{3 Clinical Diagnosis and Treatment Case Set}: Includes 35 questions from 'Pediatrics in TCM', 35 questions from 'Gynecology in TCM', 40 questions from 'Internal Medicine in TCM', and 40 questions from 'Surgery in TCM*, totaling 150 questions. This set evaluates the model's comprehensive analysis and decision-making abilities in real clinical scenarios. Its strong practicality and high comprehensiveness effectively test the model's ability to translate theoretical knowledge into practical clinical skills, making it an important basis for assessing the model's clinical application value.

\subsection{Evaluation Methods and Indicators}\label{subsec2}

\begin{figure}[H]
    \centering
    \includegraphics[width=1.0\linewidth]{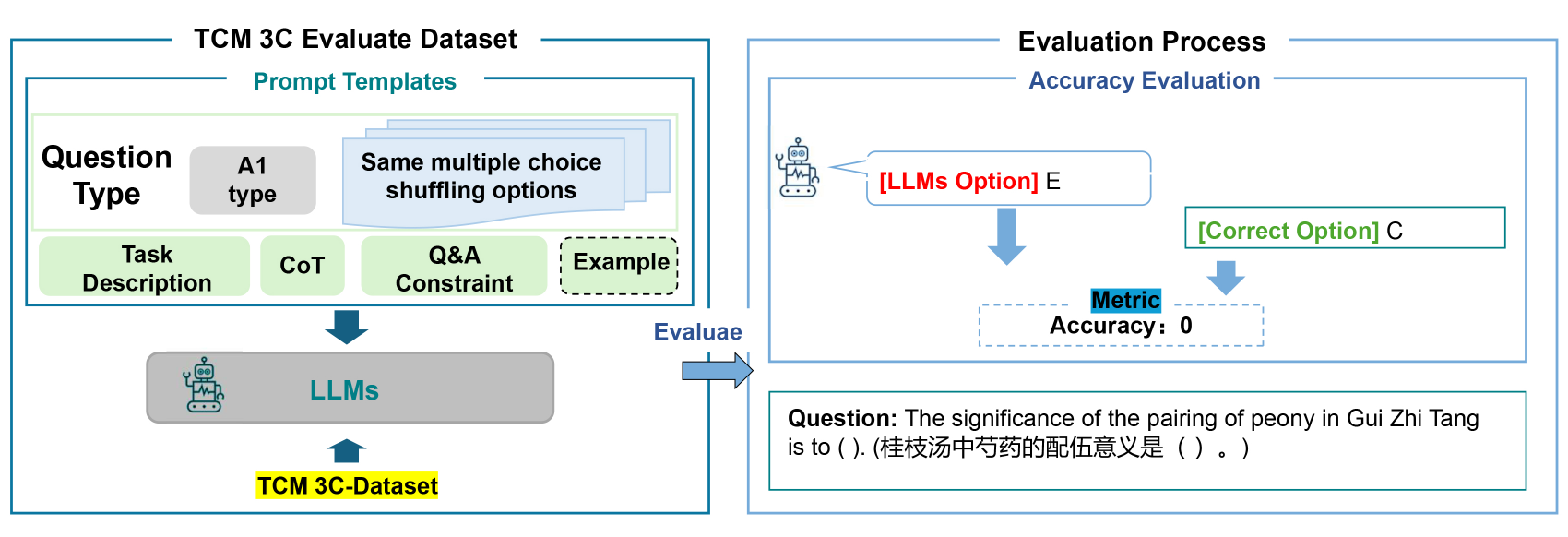}
    \caption{The overview of TCM-3CEval. It consists of two parts: (1) On the left is the construction process of the dataset. (2) On the right is the evaluation process of TCM-3CEval.} 
    \label{Figure 2}
\end{figure}

We utilize accuracy as the evaluation metric by comparing the options generated by LLMs against the correct options to assess their understanding and application of TCM knowledge. Additionally, a more robust evaluation method is employed where the options for the same multiple choice question are shuffled and presented to the models. Only if all model outputs consistently point to the same correct answer is the response considered as passed, as depicted in \autoref{Figure 2}.

\section{Results}\label{sec11}
\subsection{LLMs performance in different medical datasets.}\label{subsec2}
We conducted a comprehensive analysis of various LLMs' accuracy metrics across three types of datasets. A prominent observation reveals that DeepSeek consistently outperforms other models across all datasets, demonstrating exceptional capabilities in fundamental theories of TCM, TCM classical text comprehension, and TCM clinical case diagnosis decision-making. Models like PULSE and Internlm2.5, which underwent specialized training on extensive Chinese corpora or high-quality TCM datasets, also delivered relatively satisfactory performance. However, international models including o1-mini, gpt-4o, llama3, and claude showed suboptimal accuracy across all three types of datasets, indicating substantial room for improvement in practical clinical case applications (\autoref{Figure 3}a).

\subsection{Performance of LLMs across different medical branches.}\label{subsec2}
\begin{figure}[H]
    \centering
    \includegraphics[width=0.8\linewidth]{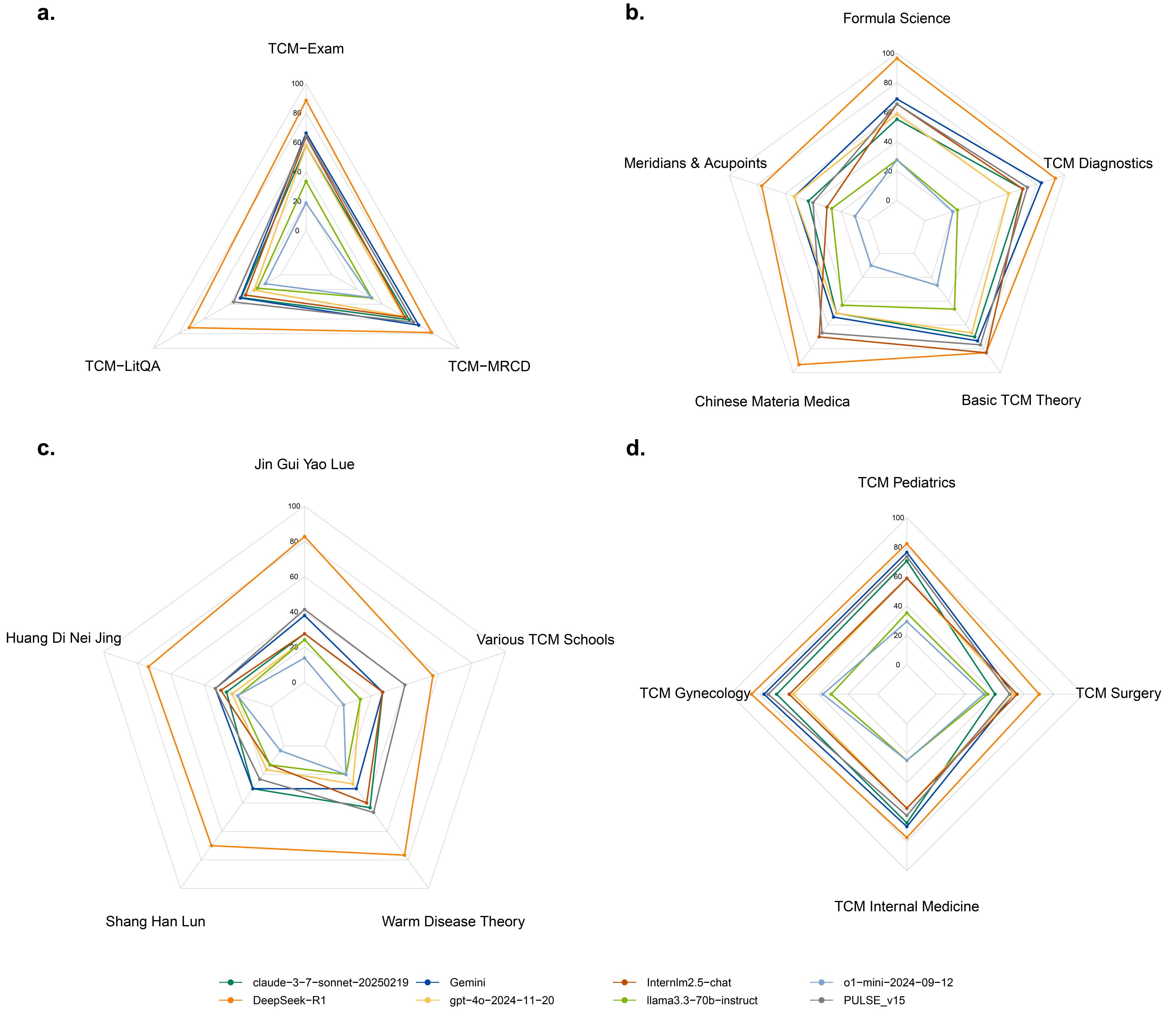}
    \caption{Benchmark performance of LLM in 3 core dimensions and sub-dimensions. (a) LLM performance on TCM-Exam, TCM-Exam, and TCM-Exam; (b) LLM performance on each branch of Basic TCM Knowledge Mastery; (c) LLM performance on each branch of Understanding of TCM Classics; (d) LLM performance on each branch of TCM Clinical Decision-Making Ability branches.}
    \label{Figure 3}
\end{figure}

We further evaluated their accuracy in addressing questions across different core dimensions. In the dimension of basic TCM knowledge mastery, DeepSeek demonstrated exceptional performance in Formula Science, Chinese Materia Medica, and Meridians \& Acupoints. Excluding DeepSeek, other models showed relatively lower accuracy in Meridians \& Acupoints compared to the other four aspects, indicating a general deficiency in this particular area. Furthermore, o1-mini and llama3 exhibited consistently poor performance across all five aspects (\autoref{Figure 3}b). Regarding the understanding of TCM classics, DeepSeek substantially outperformed other models across all five aspects, while all models showed relatively lower accuracy in Various TCM Schools compared to the other four aspects. Notably, o1-mini demonstrated particularly weak performance across all aspects, especially in Jin Gui Yao Lue, Shang Han Lun, and Various TCM Schools (\autoref{Figure 3}c). In the dimension of TCM clinical decision-making capability, DeepSeek maintained superior performance across all five aspects though with narrower margins. All models exhibited relatively lower accuracy in TCM Surgery. Both o1-mini and llama3 showed consistently poor performance across four aspects (\autoref{Figure 3}).

\subsection{Text generation length of LLMs}\label{subsec2}
\begin{figure}[H]
    \centering
    \includegraphics[width=0.8\linewidth]{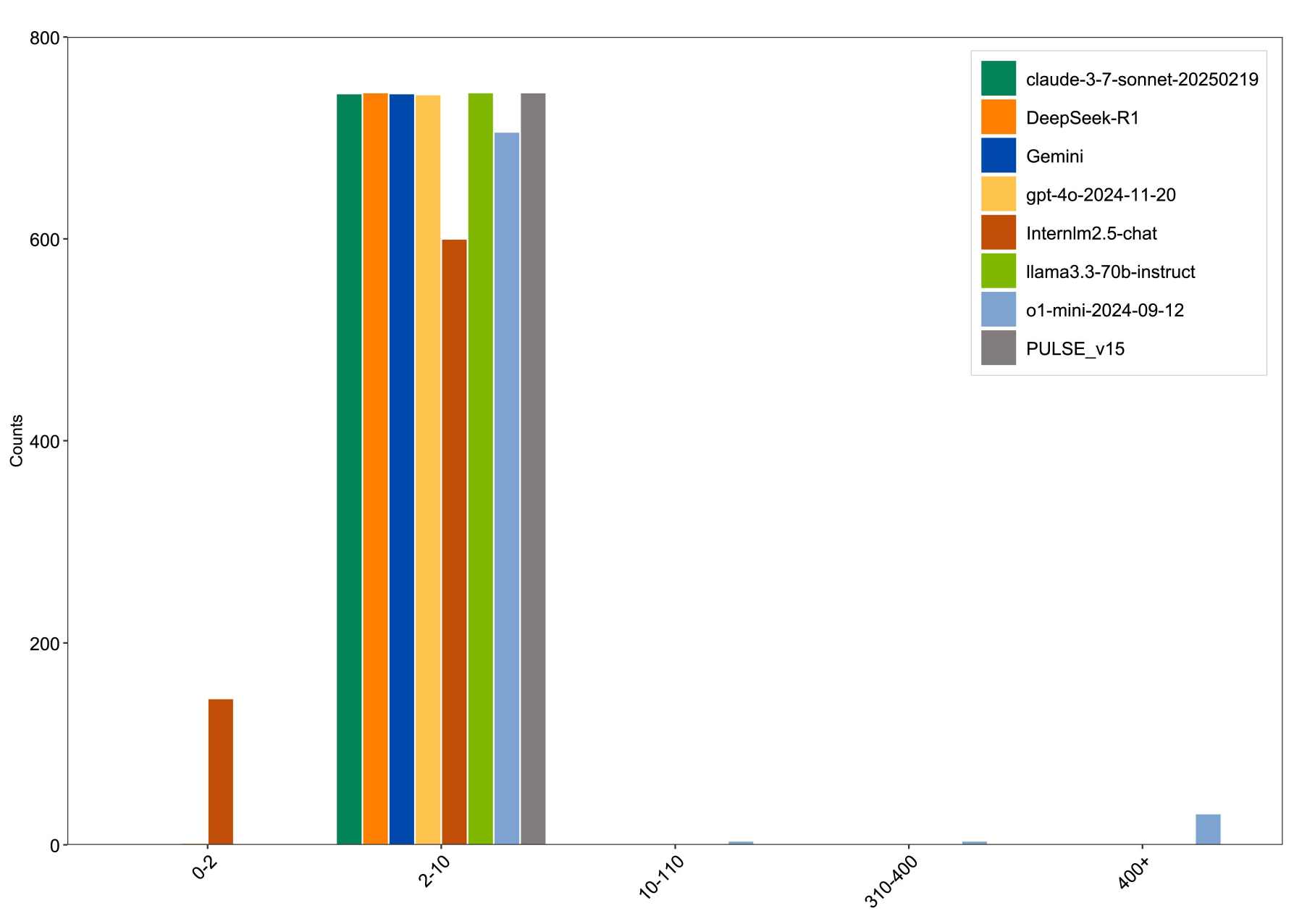}
    \caption{Text generation length of LLMs}
    \label{Figure 4}
\end{figure}

Figure \autoref{Figure 4} illustrates the text generation length across different models. Only o1-mini has appeared with 400+ generated text lengths, and all other models have generated text lengths in the range of 2-10 lengths.

\section{Discussion}\label{sec12}

\subsection{Framework Design and Evaluation Dimensions}\label{subsubsec2}

This study systematically evaluates mainstream LLMs through a framework that deliberately mirrors the established pathway of TCM physician capability development\cite{yang2023research,tao2023investigating}. The tri-dimensional structure of our evaluation dataset—Core Knowledge (TCM-Exam), Classical Literacy (TCM-LitQA), and Clinical Decision-Making (TCM-MRCD)—directly corresponds to the professional competency assessment standards used in TCM physician qualification examinations across China. This design reflects the unique educational philosophy of TCM\cite{hua2017integrating}, which differs fundamentally from modern medical training by emphasizing the integration of theoretical foundations, classical textual understanding, and pattern-based clinical reasoning\cite{chen2016comparison}. Each dimension serves as a critical evaluation point that TCM educators and licensing bodies have historically used to assess practitioner competency. Core Knowledge tests the foundational understanding required for any clinical reasoning, much as TCM medical schools evaluate students' grasp of basic theories before clinical rotations\cite{matos2021understanding}. The Classical Literacy dimension acknowledges the continued relevance of ancient medical texts as living clinical guides rather than merely historical documents—a distinctive aspect of TCM practice where physicians are expected to derive diagnostic and treatment insights directly from classical sources\cite{liu2021interpretation}. Finally, the Clinical Decision-Making dimension evaluates the practical synthesis of knowledge and classical understanding in patient-centered scenarios, mirroring the case-based assessments used in professional TCM physician examinations.

% Experimental results demonstrate DeepSeek's significant advantages in key metrics: 88.59\% accuracy in TCM-Exam, 71.81\% accuracy in TCM-LitQA, and 78.52\% accuracy in TCM-MRCD. These advantages likely stem from its dual-phase training architecture: after general Chinese corpus pre-training, domain-adaptive fine-tuning using TCM classical literature (including “Shanghan Lun” and “Jingui Yaolue”) establishes syndrome differentiation thinking patterns unique to TCM. Notably, in the TCM-LitQA task, DeepSeek achieved 71.81\% accuracy – 50 percentage points higher than general-purpose models – aligning with findings from XX et al. regarding XX. 

% The study also reveals limitations in specialized models like PULSE and Internlm2.5. While demonstrating moderate performance in structured tasks like herbal formula compatibility (average accuracy 64.10\%), they show higher error rates in classical text comprehension and clinical decision-making. This suggests that simply expanding TCM corpus size cannot fully address deep clinical reasoning requirements, necessitating integration of modern medical knowledge graphs with TCM diagnostic logic. Notably, international general models exhibited critical errors in XX, directly attributable to insufficient exposure to TCM classical texts during training – consistent with conclusions from XX et al.'s research on XX.

\subsection{Experimental Results and TCM-Specific Insights}\label{subsubsec2}

Experimental results demonstrate DeepSeek's significant advantages in key metrics: 88.59\% accuracy in TCM-Exam, 71.81\% accuracy in TCM-LitQA, and 78.52\% accuracy in TCM-MRCD. From a TCM perspective, these performance differences reflect the foundational importance of conceptual alignment with TCM's distinctive epistemological framework \cite{ding2024differences}. DeepSeek's superior performance in classical text interpretation (50 percentage points higher than general-purpose models) suggests successful internalization of TCM's unique "way of knowing" that differs fundamentally from modern medical reasoning. TCM diagnosis relies heavily on pattern recognition across seemingly unrelated symptoms, guided by theoretical constructs like Yin-Yang balance, Five Elements correspondence, and Zang-Fu organ systems functioning \cite{lu2011chinese, matos2021can}. Models lacking sufficient exposure to these conceptual frameworks struggle particularly with syndrome differentiation tasks that require holistic pattern recognition rather than linear diagnostic reasoning \cite{yu2024enhancing, ren2024large}. This explains why international models performed adequately on memorization-heavy tasks like herb properties (average 51.23\%) but faltered significantly on syndrome-based clinical reasoning tasks (average 32.18\%). The results highlight how TCM's emphasis on contextual relationships between symptoms, rather than isolated symptom-disease correlations, creates a distinctive cognitive challenge that requires specialized training to navigate effectively \cite{li2011philosophic}. Recognizing DeepSeek's superior capabilities in TCM knowledge representation, researchers have already begun applying it to specialized TCM studies, as evidenced by recent work exploring Tai Chi and Qigong applications in medical research, where DeepSeek demonstrated exceptional ability to synthesize complex traditional practices within modern research frameworks despite limitations in citation generation\cite{mcgee2025leveraging}.

\subsection{Technical Insights and Model Architecture}\label{subsubsec2}

From a technical perspective, the study reveals crucial insights about model architecture and training methodology for specialized domains like TCM. DeepSeek's dual-phase training architecture—starting \cite{wang2024reinforcement} with broad Chinese corpus pre-training followed by targeted domain-adaptive fine-tuning using classical TCM literature—establishes both linguistic and conceptual foundations necessary for TCM reasoning. This approach successfully bridges the gap between modern computational methods and TCM's classical theoretical framework, demonstrating that effective domain adaptation requires both linguistic alignment (Chinese language proficiency) and conceptual alignment (exposure to TCM-specific reasoning patterns). The limitations observed in specialized models like PULSE and Internlm2.5 provide equally valuable insights. While demonstrating moderate performance in structured tasks like herbal formula compatibility (average 64.10\%), these models showed higher error rates in classical text comprehension and clinical decision-making. This suggests that merely expanding corpus size with TCM-related texts creates factual knowledge but fails to develop the deeper clinical reasoning capabilities that characterize expert TCM practitioners. What appears missing is the integration of structured knowledge graphs that capture the complex interrelationships between TCM theoretical concepts and their clinical manifestations\cite{duan2025research}. Additionally, international general models exhibited critical errors in cultural context interpretation, including misunderstandings of polysemous Chinese medical terminology whose meanings shift based on classical text context. These findings point to the need for future models to incorporate both cultural-linguistic context and structured relationship modeling to fully capture TCM's distinctive diagnostic reasoning patterns \cite{zhu2024language}.

\subsection{TCM-3CEval Benchmark and Its Implications}\label{subsubsec2}

TCM-3CEval provides a useful framework for evaluating AI systems in the domain of Traditional Chinese Medicine \cite{zhang2023advances}. This benchmark addresses a gap in assessment tools for TCM's digital transformation, capturing the distinctive epistemological aspects of TCM—its holistic diagnostic approach, pattern-based reasoning, and integration of classical knowledge with clinical practice \cite{li2024opportunities}. By establishing standardized evaluation criteria across three dimensions (Core Knowledge, Classical Literacy, and Clinical Decision-Making), TCM-3CEval enables methodical comparison between different model architectures and training approaches, potentially accelerating improvements in TCM AI systems. The benchmark may contribute to preserving TCM's cultural heritage while supporting its modernization and international recognition \cite{zhang2022experience}. As TCM increasingly integrates with global healthcare, AI applications evaluated through this framework could help connect traditional wisdom with contemporary medical science, potentially enhancing TCM's accessibility and standardization without compromising its theoretical foundations \cite{wang2021current}. The performance differences revealed through TCM-3CEval highlight the need for specialized training approaches that respect TCM's unique conceptual foundations and linguistic contexts—insights that will be helpful for developing AI systems capable of authentic TCM reasoning rather than simple pattern matching.

\subsection{Study Limitations and Future Directions}\label{subsubsec2}

Study limitations include test sets primarily derived from literature cases, requiring future validation through real-world clinical electronic medical records. Additionally, polysemous TCM terminology\cite{hou2025pruning,Sun2024MetaphoricalTI} (e.g., contextual variations of "Qi Zhi" across classical texts) remains challenging for model comprehension. We suggest implementing a concept disambiguation mechanism with expert annotation in future research, combined with annotations of classical TCM literature and modern medical knowledge graphs, to further enhance the model's ability to understand complex TCM theories\cite{ma2022traditional}. From the perspective of TCM theory, the improvement of model performance relies not only on the increase in data volume but also on a deeper understanding of the holistic concept and syndrome differentiation and treatment principles of TCM. The core of these theories lies in emphasizing individualized diagnosis\cite{li2021enlightenment} and treatment and dynamic balance\cite{chan2022dynamic}, which poses higher demands on the model's reasoning capabilities. Future research should further explore how to integrate the holistic thinking of TCM with modern artificial intelligence technologies\cite{li2024opportunities} to promote the intelligent development of TCM.

\section*{Conclusion}
This study introduces TCM-3CEval, a pioneering benchmark for systematically evaluating LLMs' capabilities in TCM. Through our triaxial evaluation framework—encompassing Core Knowledge mastery, Classical Literacy, and Clinical Decision-Making—we have identified critical patterns in model performance that offer valuable insights for future development. Our findings clearly demonstrate that domain specialization significantly impacts TCM competency, with models specifically trained on TCM corpora substantially outperforming general-purpose models. The remarkable performance gap between culturally-aligned models like DeepSeek and international counterparts underscores the essential role of cultural-linguistic context in representing TCM's unique epistemological framework. However, even the best-performing models exhibit notable deficiencies in specialized areas such as Meridian \& Acupoint theory and Various TCM Schools, revealing persistent gaps between current AI capabilities and the comprehensive expertise required for clinical practice. These limitations point to the need for more sophisticated integration of TCM's holistic diagnostic paradigm with modern computational approaches—beyond mere corpus expansion to include structured knowledge representation capturing the complex interrelationships between theoretical concepts and clinical manifestations. TCM-3CEval establishes a foundation for future research directions, including the development of multi-modal evaluation incorporating visual diagnostic elements (tongue and pulse diagnosis), integration of real-world clinical data beyond textbook cases, and exploration of how LLMs might support the standardization and internationalization of TCM while preserving its distinctive theoretical framework. By providing a standardized, comprehensive assessment methodology, this benchmark contributes to the advancement of TCM-oriented AI systems that can effectively bridge traditional wisdom with modern technology.

\section*{Acknowledgments}
This work was supported by the National Natural Science Foundation of
China [grant 82174506] and the Traditional Chinese Medicine Research Project of Shanghai Municipal Health Commission [grant 2024PT001].

\section*{Declarations}
The authors declare that the research was conducted in the absence of any commercial or financial relationships that could be construed as a potential conflict of interest.

%Bibliography
\bibliographystyle{unsrt}  
\bibliography{references}  

\end{document}